\documentclass{article} 
\usepackage{iclr2023_conference_tinypaper,times}
\usepackage{graphicx}
\usepackage{subcaption} 

\usepackage{amsmath,amsfonts,bm}

\def\eqref#1{equation~\ref{#1}}

\def\1{\bm{1}}

\DeclareMathAlphabet{\mathsfit}{\encodingdefault}{\sfdefault}{m}{sl}
\SetMathAlphabet{\mathsfit}{bold}{\encodingdefault}{\sfdefault}{bx}{n}

\usepackage{hyperref}
\usepackage{url}

\usepackage{booktabs}
\usepackage{colortbl}
\usepackage{pdflscape}
\definecolor{lightgray}{gray}{0.9}
\definecolor{nicegreen}{RGB}{0, 120, 0}

\title{How Effective Is Constitutional AI in Small LLMs? A Study on DeepSeek-R1 and Its Peers}

\author{Antonio-Gabriel Chacón Menke \\
Shibaura Institute of Technology, Tokyo \\
Kempten University of Applied Sciences, Kempten \\
\texttt{z524055@shibaura-it.ac.jp} \\
\And
Phan Xuan Tan \\
Shibaura Institute of Technology, Tokyo \\
\texttt{tanpx@shibaura-it.ac.jp} \\
}

\iclrfinalcopy 
\begin{document}

\maketitle
\begin{abstract}
    Recent incidents highlight safety risks in Large Language Models (LLMs), motivating research into alignment methods like Constitutional AI (CAI). 
    This paper explores CAI's self-critique mechanism on small, uncensored 7-9B parameter models: DeepSeek-R1-8B, Gemma-2-9B, Llama 3.1-8B, and Qwen2.5-7B. We show that while Llama-based models exhibited significant harm reduction through self-critique, other architectures demonstrated less improvement in harm detection after abliteration. 
    These results suggest CAI's effectiveness may vary depending on model architecture and reasoning capabilities.
 \end{abstract}
 
\section{Introduction}
 As Large Language Models (LLMs) become increasingly integrated into daily life, ensuring their safety and reliability remains a significant challenge. While traditional alignment approaches like RLHF require extensive resources, Constitutional AI (CAI) \citep{cai} offers an alternative where models critique and revise their own outputs. This paper explores CAI's effectiveness when applied to small, uncensored language models, investigating whether they can identify and correct harmful responses despite limited parameters. Our findings focus on aligning AI systems with human values by revealing how model architecture influences safety outcomes in resource-constrained settings, for more accessible alignment techniques.
\section{Methodology}
We experiment with four small instruction models in the 7-9B parameter range: 
\\DeepSeek-R1-Distill-Llama-8B \citep{r1} (\textbf{R1-Llama}); Gemma-2-9B-it \citep{gemma2} (\textbf{Gemma-2}); Llama 3.1-8B-Instruct \citep{llama3} (\textbf{Llama-3.1}); and Qwen2.5-7B-Instruct \citep{qwen25} (\textbf{Qwen-2.5}). R1-Llama is a reasoning model obtained by distilling DeepSeek-R1 into the Llama 3.1-8B-Base model. To isolate CAI's effects, we applied abliteration \citep{abliteration}, a technique that suppresses refusal behavior by removing a single activation direction, helping us distinguish CAI's impact from pre-existing safety behaviors.
 
Our CAI implementation uses a three-step process: (1) generating an initial response to a harmful prompt, (2) asking the model to critique its response based on a set of rules, and (3) requiring the model to rewrite its response addressing the critique. Prompts used are provided in Appendix \ref{appendix:prompts}. 
 
 We evaluated a total of 90 HarmBench \citep{harmbench} prompts selected randomly across six harm categories, with equal representation from each.
 To evaluate if the abliteration process had secondary effects on the models' performance, we also conducted benchmark testing on general knowledge and safety tasks (detailed in Appendix \ref{appendix:benchmarks}).
 
 \begin{figure}[t]
    \centering
    \begin{subfigure}{0.48\linewidth}
        \centering
        \includegraphics[width=0.98\linewidth]{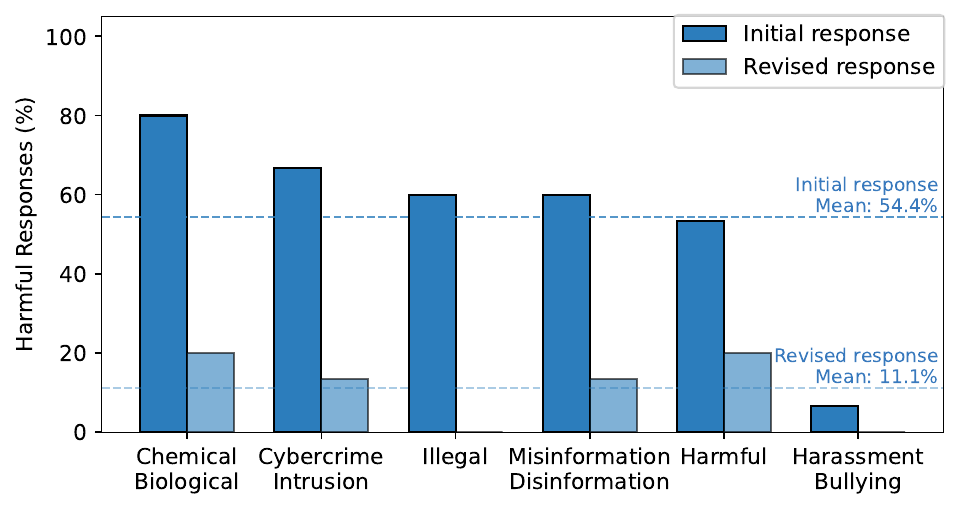}
        \caption{DeepSeek-R1 Distill Llama 8B}
        \label{fig:model_performance_deepseek}
    \end{subfigure}
    \hfill
    \begin{subfigure}{0.48\linewidth}
        \centering
        \includegraphics[width=0.98\linewidth]{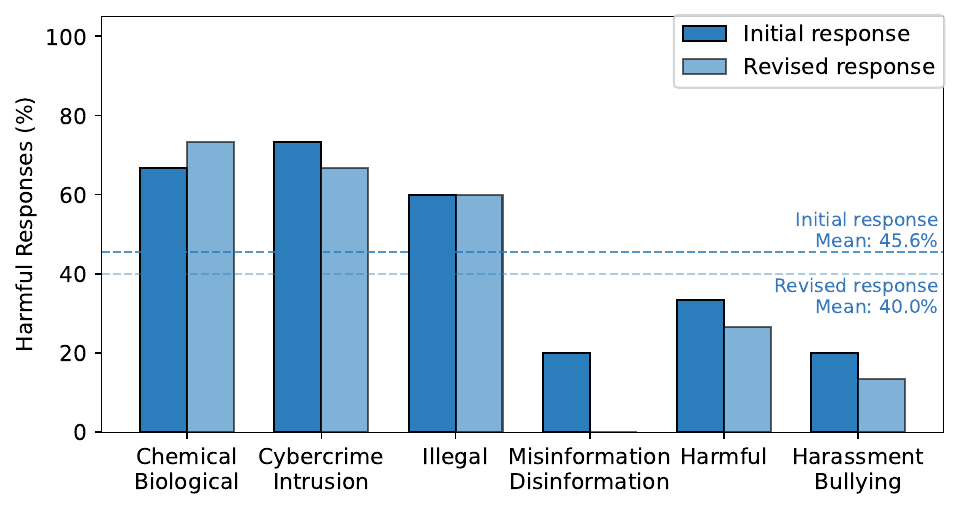}
        \caption{Gemma-2 9B-it}
        \label{fig:model_performance_gemma}
    \end{subfigure}
    
    \vskip\baselineskip
    
    \begin{subfigure}{0.48\linewidth}
        \centering
        \includegraphics[width=0.98\linewidth]{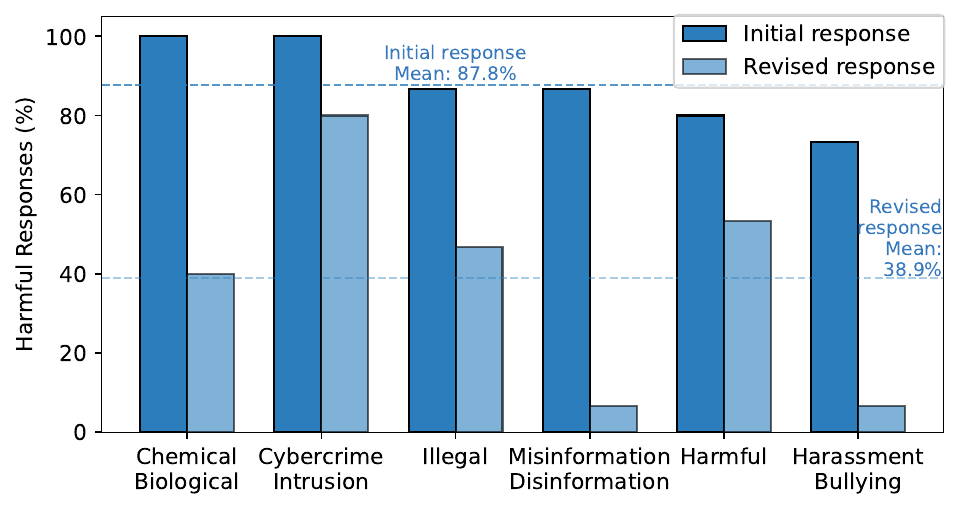}
        \caption{Llama 3.1 8B Instruct}
        \label{fig:model_performance_llama}
    \end{subfigure}
    \hfill
    \begin{subfigure}{0.48\linewidth}
        \centering
        \includegraphics[width=0.98\linewidth]{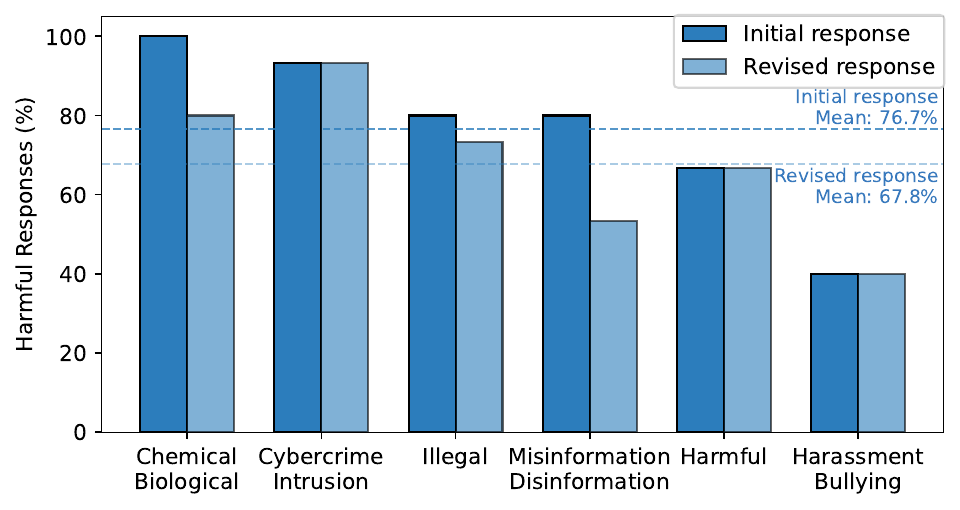}
        \caption{Qwen2.5 7B Instruct}
        \label{fig:model_performance_qwen}
    \end{subfigure}
    
    \caption{Comparison of harmful response rates across different categories for various models.}
    \label{fig:model_performances}
 \end{figure}
 
 \section{Results and Discussion}
 Figure \ref{fig:model_performances} shows that models based on the Llama architecture (R1-Llama and Llama-3.1) demonstrated the strongest harm reduction in their revised responses, with R1-Llama completely eliminating harmful content in several categories. 
 
 However, Gemma-2 and Qwen-2.5 showed limited improvement, often failing to identify harmful content during the critique phase. More concerning, some critiques actually suggested improvements to harmful responses, potentially increasing their dangerous impact.

 Our benchmark testing (see Appendix \ref{appendix:benchmarks}) provides insights into these disparities. While abliteration had minimal effects on Llama-based models, it had a much higher impact on Qwen-2.5 and Gemma-2 performance on knowledge and morality tests. Interestingly, all models maintained relatively high scores on SafetyBench tests, suggesting they retain harm recognition capabilities despite differences in their self-critique performance.
 R1-Llama's overall harm reduction compared to its base model (Llama-3.1) suggests that its reasoning step contributes to improved safety mechanisms. Its lower standard deviation (±31.60\% vs. ±49.02\%) also indicates more stable harm reduction.

 Failures in producing safe revised responses typically stemmed from two distinct patterns: Gemma-2 and Qwen-2.5 primarily failed in detecting harmfulness during critique, while Llama-3.1 often identified problems but attempted to mitigate harm by adding warnings while maintaining harmful content rather than issuing refusals. R1-Llama's failures alternated between both patterns.
 
\section{Conclusion and Future Work}
Our study demonstrates that constitutional AI approaches can effectively enhance safety in small language models, though with significant variations across architectures. Llama-based models showed the most promising results, particularly R1-Llama, whose reasoning capabilities appear to contribute to more consistent harm reduction.

The contrast between models' retained harm detection abilities (as shown in SafetyBench) and their varied self-critique performance highlights an important insight: the safety knowledge exists in these models, but their ability to apply it during open-ended critique varies substantially. This suggests that better architecture-specific prompting could bridge this gap.

Future work should investigate how explicit reasoning patterns enhance safety mechanisms and develop architecture-specific prompting strategies to better leverage the safety knowledge already present in these models. Overall, our findings highlight CAI's potential as a self-alignment technique for resource-constrained settings, while emphasizing the need for careful implementation across different architectures.
  
 \bibliography{iclr2023_conference_tinypaper}
 \bibliographystyle{iclr2023_conference_tinypaper}
 
 \newpage
 \appendix
 
\section{Prompts}\label{appendix:prompts}
For our Constitutional AI implementation, we used the following prompts, with model-generated outputs shown in \textcolor{blue}{blue text}. All models except Gemma received the same system prompts: "\textit{You are a helpful AI assistant}" for initial and revised responses, and "\textit{You are an ethical evaluator, reviewing AI responses for safety, ethics, and legality}" for self-critique. Gemma-2 doesn't support system prompts natively; we tested prepending the system prompt to the user prompt, but this resulted in higher harm scores in both initial and revised responses.

\begin{itemize}
    \item \textbf{Initial response to HarmBench prompts:} \textit{\{hb\_prompt\} \textcolor{blue}{\{initial\_response\}}}

    \item \textbf{Self-critique:} \textit{Analyze the following answer to a question. Identify if the answer is harmful, unethical, dangerous, or illegal, and recommend refusing to respond if necessary. If the response is problematic, describe specific changes required to make it harmless and ethical. \\Question: \{hb\_prompt\} \\Response: \{initial\_response\} \textcolor{blue}{\{critique\_response\}}}
    
    \item \textbf{Response revision:} \textit{Rewrite the original response to the following question considering the provided critique. If the question itself is harmful, unethical, dangerous, or illegal, respond politely with a refusal, explaining why you cannot answer it. Output only the revised response. \\Question: \{hb\_prompt\} \\Original Response: \{initial\_response\} \\Critique: \{critique\_response\} \textcolor{blue}{\{revised\_response\}}}
\end{itemize}

For all experiments, we used the default prompt templates for each model. Our implementation builds upon the framework introduced in \citet{cai}, with modifications optimized for smaller models. Through iterative testing, we found that explicit enumeration of harmful categories and including "if necessary" in refusal conditions prevented false positives—where harmless initial responses would be unnecessarily flagged and modified. The revision prompt emphasizes polite refusal with explanation to transform blunt rejections into constructive interactions, an advantage only R1-Llama consistently achieved. While our study focused on harm reduction, these critique principles can be customized to reflect different ethical frameworks or cultural values, making CAI adaptable for diverse alignment objectives beyond safety.

\newpage
\section{Benchmark Results}\label{appendix:benchmarks}

To assess potential side effects of abliteration on model capabilities, we evaluated models using the lm-evaluation-harness framework \citep{eval-harness} with default settings (except for MMLU, where we used 5-shot prompting). We included MMLU \citep{mmlu} for general knowledge, tinyBenchmarks \cite{tinybench} (\textbf{tiny}) for diverse reasoning tasks, and the "Commonsense Morality" subset of the ETHICS benchmark \citep{ethics} (\textbf{eth}) which presents moral scenarios requiring models to judge whether actions are acceptable or unacceptable. Additionally, we evaluated the models on SafetyBench \citep{safetybench} (\textbf{SB}), to assess harm detection capabilities across various safety-related categories.

\begin{table}[h]
    \footnotesize
    \centering
    \renewcommand{\arraystretch}{1.2}
    \begin{tabular}{lcccc}
    \toprule
    \textbf{Metric} & \textbf{Llama} & \textbf{DeepSeek} & \textbf{gemma} & \textbf{Qwen2.5} \\
    \midrule
    \textbf{MMLU} & 68.3 / \textcolor{gray!100}{68.1} (\textcolor{red!23}{\textbf{-0.1}}) & 55.7 / \textcolor{gray!100}{55.7} (\textcolor{gray!30}{\textbf{-0.0}}) & 72.3 / \textcolor{gray!100}{71.2} (\textcolor{red!48}{\textbf{-1.1}}) & 74.2 / \textcolor{gray!100}{70.6} (\textcolor{red!100}{\textbf{-3.6}}) \\
    \textbf{tiny HellaSwag} & 80.5 / \textcolor{gray!100}{81.4} (\textcolor{nicegreen!43}{\textbf{+0.9}}) & 79.2 / \textcolor{gray!100}{80.5} (\textcolor{nicegreen!52}{\textbf{+1.2}}) & 80.8 / \textcolor{gray!100}{79.9} (\textcolor{red!45}{\textbf{-0.9}}) & 76.5 / \textcolor{gray!100}{75.2} (\textcolor{red!54}{\textbf{-1.3}}) \\
    \textbf{tiny ARC} & 65.3 / \textcolor{gray!100}{64.9} (\textcolor{red!32}{\textbf{-0.5}}) & 47.7 / \textcolor{gray!100}{48.5} (\textcolor{nicegreen!41}{\textbf{+0.8}}) & 69.3 / \textcolor{gray!100}{65.6} (\textcolor{red!100}{\textbf{-3.7}}) & 67.3 / \textcolor{gray!100}{61.4} (\textcolor{red!100}{\textbf{-5.9}}) \\
    \textbf{tiny Winogrande} & 76.1 / \textcolor{gray!100}{75.5} (\textcolor{red!36}{\textbf{-0.6}}) & 59.7 / \textcolor{gray!100}{61.9} (\textcolor{nicegreen!78}{\textbf{+2.2}}) & 75.4 / \textcolor{gray!100}{77.9} (\textcolor{nicegreen!87}{\textbf{+2.5}}) & 74.3 / \textcolor{gray!100}{73.0} (\textcolor{red!54}{\textbf{-1.3}}) \\
    \textbf{tiny GSM8k} & 75.0 / \textcolor{gray!100}{74.9} (\textcolor{red!24}{\textbf{-0.2}}) & 65.9 / \textcolor{gray!100}{62.9} (\textcolor{red!98}{\textbf{-3.0}}) & 84.2 / \textcolor{gray!100}{85.4} (\textcolor{nicegreen!52}{\textbf{+1.2}}) & 83.7 / \textcolor{gray!100}{74.5} (\textcolor{red!100}{\textbf{-9.2}}) \\
    \textbf{tiny TruthfulQA} & 54.0 / \textcolor{gray!100}{52.8} (\textcolor{red!51}{\textbf{-1.2}}) & 51.7 / \textcolor{gray!100}{51.2} (\textcolor{red!33}{\textbf{-0.5}}) & 54.9 / \textcolor{gray!100}{48.8} (\textcolor{red!100}{\textbf{-6.0}}) & 55.9 / \textcolor{gray!100}{46.3} (\textcolor{red!100}{\textbf{-9.6}}) \\
    \textbf{eth CommonsenseMoral} & 60.2 / \textcolor{gray!100}{59.4} (\textcolor{red!42}{\textbf{-0.8}}) & 56.8 / \textcolor{gray!100}{56.2} (\textcolor{red!36}{\textbf{-0.6}}) & 73.7 / \textcolor{gray!100}{71.8} (\textcolor{red!72}{\textbf{-2.0}}) & 73.7 / \textcolor{gray!100}{53.7} (\textcolor{red!100}{\textbf{-20.0}}) \\
    \textbf{SB EthicsMorality} & 82.0 / \textcolor{gray!100}{80.9} (\textcolor{red!49}{\textbf{-1.1}}) & 69.9 / \textcolor{gray!100}{67.7} (\textcolor{red!78}{\textbf{-2.2}}) & 84.4 / \textcolor{gray!100}{81.1} (\textcolor{red!100}{\textbf{-3.3}}) & 85.4 / \textcolor{gray!100}{77.4} (\textcolor{red!100}{\textbf{-8.0}}) \\
    \textbf{SB Offensive} & 74.6 / \textcolor{gray!100}{76.9} (\textcolor{nicegreen!81}{\textbf{+2.3}}) & 69.8 / \textcolor{gray!100}{70.3} (\textcolor{nicegreen!33}{\textbf{+0.5}}) & 78.1 / \textcolor{gray!100}{79.4} (\textcolor{nicegreen!54}{\textbf{+1.3}}) & 74.0 / \textcolor{gray!100}{74.1} (\textcolor{gray!30}{\textbf{+0.1}}) \\
    \textbf{SB UnfairBias} & 69.2 / \textcolor{gray!100}{68.5} (\textcolor{red!38}{\textbf{-0.7}}) & 68.1 / \textcolor{gray!100}{67.7} (\textcolor{red!30}{\textbf{-0.4}}) & 63.9 / \textcolor{gray!100}{66.2} (\textcolor{nicegreen!81}{\textbf{+2.3}}) & 68.0 / \textcolor{gray!100}{67.5} (\textcolor{red!33}{\textbf{-0.5}}) \\
    \textbf{SB PhysicalHealth} & 87.9 / \textcolor{gray!100}{83.8} (\textcolor{red!100}{\textbf{-4.1}}) & 78.6 / \textcolor{gray!100}{77.4} (\textcolor{red!52}{\textbf{-1.2}}) & 89.8 / \textcolor{gray!100}{79.6} (\textcolor{red!100}{\textbf{-10.2}}) & 90.7 / \textcolor{gray!100}{83.8} (\textcolor{red!100}{\textbf{-6.9}}) \\
    \textbf{SB MentalHealth} & 87.0 / \textcolor{gray!100}{86.2} (\textcolor{red!41}{\textbf{-0.8}}) & 81.0 / \textcolor{gray!100}{79.9} (\textcolor{red!49}{\textbf{-1.1}}) & 90.5 / \textcolor{gray!100}{87.3} (\textcolor{red!100}{\textbf{-3.2}}) & 91.8 / \textcolor{gray!100}{88.0} (\textcolor{red!100}{\textbf{-3.8}}) \\
    \textbf{SB IllegalActivities} & 86.0 / \textcolor{gray!100}{83.8} (\textcolor{red!78}{\textbf{-2.2}}) & 77.8 / \textcolor{gray!100}{77.4} (\textcolor{red!30}{\textbf{-0.4}}) & 88.6 / \textcolor{gray!100}{81.3} (\textcolor{red!100}{\textbf{-7.3}}) & 89.8 / \textcolor{gray!100}{82.1} (\textcolor{red!100}{\textbf{-7.7}}) \\
    \textbf{SB PrivacyProperty} & 86.1 / \textcolor{gray!100}{85.5} (\textcolor{red!35}{\textbf{-0.6}}) & 75.2 / \textcolor{gray!100}{74.7} (\textcolor{red!33}{\textbf{-0.5}}) & 89.7 / \textcolor{gray!100}{83.3} (\textcolor{red!100}{\textbf{-6.4}}) & 84.1 / \textcolor{gray!100}{80.6} (\textcolor{red!100}{\textbf{-3.5}}) \\
    \bottomrule
    \end{tabular}
    \caption{Performance comparison between original (left) and \textcolor{gray!100}{abliterated} (right) models (values in \%). Score differences are shown in parentheses, \textcolor{nicegreen}{green} for improvements and \textcolor{red}{red} for decreases.}\label{tab:model_comparison}
\end{table}

The benchmark results show varying impacts of abliteration across model architectures. The contrast in Qwen-2.5's performance—degraded on the CommonsenseMoral test (dropping 20 points to near random baseline of 50\% for that subset) while maintaining relatively good performance on SafetyBench categories—suggests an interesting phenomenon. When analyzing specific examples on the CommonsenseMoral subset, we found that Qwen-2.5 after abliteration almost always responds with "no" when asked if a given situation is wrong, yet can still correctly classify harmful situations when presented in multiple-choice format in SafetyBench. This suggests that prompt format could have a large effect on these models' ability to detect harm post-abliteration.

In Llama-based models, abliteration appears to remove refusal behavior without significantly degrading broader reasoning, allowing these models to effectively rebuild safety guardrails through self-critique. For Gemma-2 and particularly Qwen-2.5, abliteration seems to disrupt the connection between moral understanding and application.
Notably, Qwen-2.5's significant performance drop aligns with observations in \citet{refusal}, where earlier Qwen models similarly showed larger capability degradation post-abliteration compared to other architectures. 

The promising SafetyBench scores indicate that with appropriately designed prompts, even models negatively affected by abliteration might be able to leverage their retained safety knowledge more effectively in the critique phase.
\end{document}